\documentclass[letterpaper]{article} 
\usepackage{aaai2026}  
\usepackage{times}  
\usepackage{helvet}  
\usepackage{courier}  
\usepackage[hyphens]{url}  
\usepackage{graphicx} 
\usepackage{natbib}  
\usepackage{caption} 
\usepackage{algorithm}
\usepackage{algorithmic}
\usepackage{booktabs}
\usepackage{amsmath} 
\usepackage{amssymb} 
\usepackage{array} 
\usepackage{multirow}
\usepackage{newfloat}
\usepackage{listings}
\DeclareCaptionStyle{ruled}{labelfont=normalfont,labelsep=colon,strut=off} 
\floatstyle{ruled}
\newfloat{listing}{tb}{lst}{}
\floatname{listing}{Listing}
\title{Explainable Melanoma Diagnosis with Contrastive Learning and LLM-based Report Generation}
\newcommand{\sharedthanks}[1]{\footnotemark[\value{footnote}]\footnotetext{#1}}
\author{
    Junwen Zheng\textsuperscript{\rm 1},
    Xinran Xu\textsuperscript{\rm 1},
    Li Rong Wang\textsuperscript{\rm 1,2},
    Chang Cai\textsuperscript{\rm 1},
    Lucinda Siyun Tan\textsuperscript{\rm 3},
    Dingyuan Wang\textsuperscript{\rm 3},
    Hong Liang Tey\textsuperscript{\rm 1,3}\thanks{Corresponding authors},
    Xiuyi Fan\textsuperscript{\rm 1}\sharedthanks{}
}
\affiliations{
    \textsuperscript{\rm 1}Nanyang Technological University, Singapore\\
    \textsuperscript{\rm 2}Centre for Frontier AI Research, A*STAR, Singapore\\
    \textsuperscript{\rm 3}National Skin Centre,
    National Healthcare Group, Singapore\\
    \{JUNWEN003, XINRAN007, LIRONG002, CHANG023\}@e.ntu.edu.sg,\\
    \{lucindatan, dywang\}@nsc.com.sg, 
    \{teyhongliang, xyfan\}@ntu.edu.sg
}

\usepackage{bibentry}

\begin{document}

\maketitle

\begin{abstract}
Deep learning has demonstrated expert-level performance in melanoma classification, positioning it as a powerful tool in clinical dermatology. However, model opacity and the lack of interpretability remain critical barriers to clinical adoption, as clinicians often struggle to trust the decision-making processes of black-box models. To address this gap, we present a Cross-modal Explainable Framework for Melanoma (CEFM) that leverages contrastive learning as the core mechanism for achieving interpretability. Specifically, CEFM maps clinical criteria for melanoma diagnosis—namely Asymmetry, Border, and Color (ABC)—into the Vision Transformer embedding space using dual projection heads, thereby aligning clinical semantics with visual features. The aligned representations are subsequently translated into structured textual explanations via natural language generation, creating a transparent link between raw image data and clinical interpretation. Experiments on public datasets demonstrate 92.79\% accuracy and an AUC of 0.961, along with significant improvements across multiple interpretability metrics. Qualitative analyses further show that the spatial arrangement of the learned embeddings aligns with clinicians’ application of the ABC rule, effectively bridging the gap between high-performance classification and clinical trust.

\end{abstract}
\begin{links}
    \link{Code}{https://eattt-wen.github.io/CEFM/}
\end{links}

\section{Introduction}
\label{sec:introduction}
The global incidence of malignant melanoma continues to rise steadily, with this aggressive skin cancer accounting for over 80\% of skin cancer-related deaths~\cite{mallardo2025advances}. Traditional diagnosis predominantly relies on physicians' visual assessments and dermoscopic inspection. To support systematic evaluation, clinicians often refer to the well-established {\bf ABCD} rule, which assesses key dermoscopic features: {\em \underline{A}symmetry, \underline{B}order irregularity, \underline{C}olor variation, \underline{D}ifferential structure}. However, the accuracy of such evaluations is highly dependent on the clinician’s expertise and is susceptible to subjective variability.

Recent advances in convolutional neural networks (CNNs) have led to remarkable improvements in melanoma classification performance~\cite{almufareh2024melanoma,hussien2024classification}. Despite achieving high diagnostic accuracy ($\geq$90\%), most CNN-based systems suffer from a lack of interpretability, which limits their adoption in clinical practice. These models are often regarded as “black boxes,” with opaque decision-making processes that undermine clinicians’ trust in AI-assisted diagnosis~\cite{xu2024medical}. Enhancing transparency through clinically aligned interpretable representations is essential for building physician trust and integrating AI into clinical workflows~\cite{hossain2025explainable}. In this context, explainable AI (XAI) approaches that generate clinically meaningful, context-aware explanations are increasingly recognized as critical to bridging the gap between algorithmic performance and real-world clinical utility~\cite{toh2025effect}.


Existing XAI methods, such as Grad‑CAM~\cite{selvaraju2017grad}, highlight regions of interest but fail to integrate domain‑specific ABCD diagnostic priors, limiting their clinical depth and trustworthiness.

\begin{figure}
    \centering
    \includegraphics[trim={0 8.636cm 16.2306cm 0},clip, width=\linewidth]{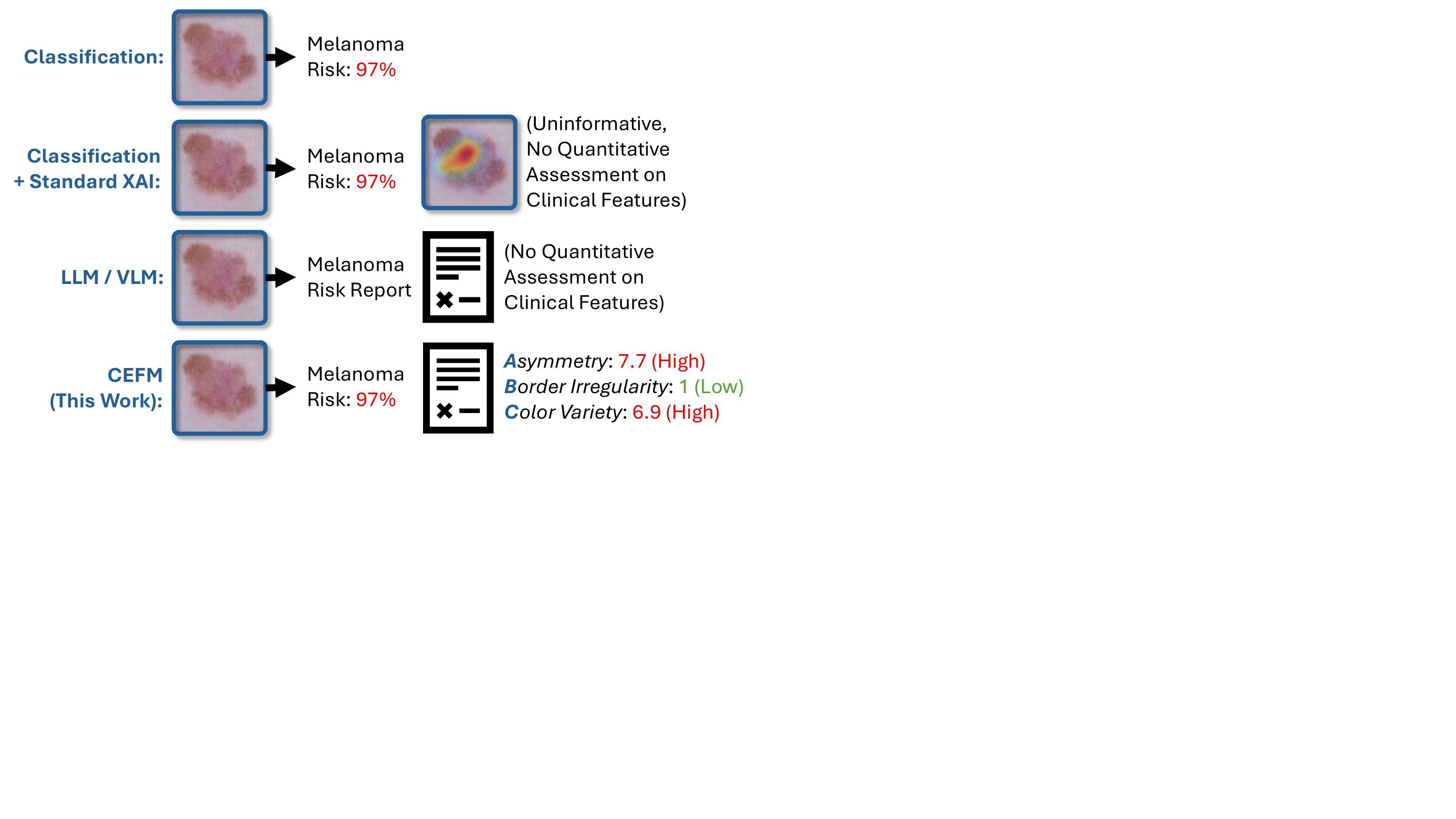}
    \caption{Advantages of the Cross-modal Explainable Framework (CEFM) over existing frameworks.}
    \label{fig:introduction_fig}
\end{figure}

To bridge this trust gap between black-box AI models and clinical deployment, we propose a {\bf Cross-modal Explainable Framework for Melanoma diagnosis (CEFM; Figure~\ref{fig:introduction_fig})}. Our system is designed to produce semantically rich explanations aligned with established diagnostic criteria, thereby enhancing the interpretability, credibility, and usability of AI in dermatological settings. The core contributions of this work are as follows:

\begin{enumerate}
\item
{\bf Clinically-Grounded Feature Extraction Pipeline:} We introduce a coarse-to-fine segmentation module that leverages UltraLight VM-UNet and SAM2 to extract three key dermoscopic features—{\bf Asymmetry, Border irregularity}, and {\bf Color variation}—based on the ABCD rule, enabling interpretable lesion characterization.

\item
{\bf Cross-Modal Contrastive Alignment:} We develop a contrastive learning architecture that aligns visual representations from a Vision Transformer with clinically features, enabling the classifier to learn semantically meaningful embeddings that reflect dermatological reasoning.

\item
{\bf Hybrid Diagnostic Report Generation:} We propose a novel explanation module integrating CLIP-based medical concept activation with a domain-adapted language model (DeepSeek) to generate structured, clinically accurate diagnostic reports supporting clinical decision-making.

\end{enumerate}
Our ablation study shows that each module in the explanation pipeline contributes uniquely to generating coherent and clinically meaningful melanoma explanations, further supported by a structured user study with three board-certified dermatologists. Experts highlighted the framework's strong interpretability (avg. score: 4.60/5), endorsed the ABC feature analysis, and affirmed its clinical utility for early triage, junior clinicians, and longitudinal lesion tracking. Furthermore, our framework demonstrates strong quantitative performance across segmentation and classification tasks. 

\section{Related Work}
\subsection{Automatic Classification of Melanoma}
In recent years, deep learning has made remarkable progress in the Automatic Classification of Melanoma in dermoscopic images. \citet{sabir2024classification} compared a baseline CNN, ResNet-18, and EfficientNet-B0, and demonstrated that EfficientNet-B0 performed best, achieving a remarkable accuracy of 97\%. In our work, we reproduce their comparison of convolutional backbones and extend it with a Vision Transformer (ViT), which outperformed EfficientNet-B0. In parallel, \citet{abir2024deep} focused on image preprocessing and feature engineering, applying noise removal and lesion segmentation strategies that significantly improved ResNet-based performance. 

 
\subsection{Multimodal Learning in Dermatological Diagnosis}
Beyond image-based unimodal models, multimodal learning approaches have gained increasing attention in dermatological diagnosis due to their ability to integrate complementary clinical information and improve diagnostic accuracy. Previous studies performed this integration using traditional algorithms such as Random Forest, Logistic Regression, and Support Vector Machines for multi-type skin disease classification and risk prediction~\cite{almustafa2025predictive}. However, these works predominantly focus on classification alone, without addressing interpretability.
\subsection{XAI for Melanoma Detection}
Existing work aimed at improving interpretability with explainable AI (XAI) includes \citet{thomas2021interpretable}, who applied a U-Net with a ResNet-50 encoder–decoder to segment skin‐tissue images into 12 semantic classes, producing tissue‐distribution maps that enable pathology-style explanations. By contrast, post-hoc saliency methods such as Grad-CAM merely highlight regions of model attention without linking them to diagnostic structures \cite{gamage2024melanoma}. Another line of work generates clinically relevant textual descriptions using LSTMs based on visual attention maps after classification~\cite{zhang2019pathologist}. However, these approaches still do not generate clinically meaningful explanations grounded in established diagnostic criteria.



In the area of Clinical Explanation of Melanoma Classification, the work by ~\citet{chanda2024dermatologist} is most closely related to ours. They introduced a guided attention mechanism (CompA), which aligns model-generated feature maps with clinician-annotated regions for more precise focus, and combines it with Grad-CAM to provide interpretable diagnoses. However, their approach requires extensive manual annotation for model construction, presenting scalability challenges. Additionally, highlighting model attention regions is perceived as less useful for clinical decision-making than textual explanations~\cite{kayser2024fool}, driving interest in explanation methods beyond visual attention maps.

\subsection{LLM‑Based Report Generation}
One strategy to producing textual explanations is to employ large language models (LLM) to generate structured reports. \citet{wang2023r2gengpt} proposed R2GenGPT, a lightweight framework that maps image features into the embedding space of a frozen LLM via a small projection module, enabling efficient end-to-end generation. Meanwhile, other studies proposed a prompt-guided framework incorporating anatomical priors and clinical context~\cite{li2024prompt}. Key anatomical regions are detected via object detection, and region-level descriptions are generated by a transformer decoder. Structured reports are then generated by the LLM with both the region-level descriptions and the clinical prompts. While this improves report structure and interactivity, the approach offers limited diagnostic guidance for generating region-level descriptions. Additionally, the clinical prompts, based on history, indications, and reasons for examination, do not explicitly reflect diagnostic criteria. As a result, the generated reports may fall short in supporting accurate diagnosis.

\begin{figure*}[!t]
    \centering
    \includegraphics[width=0.84\linewidth]{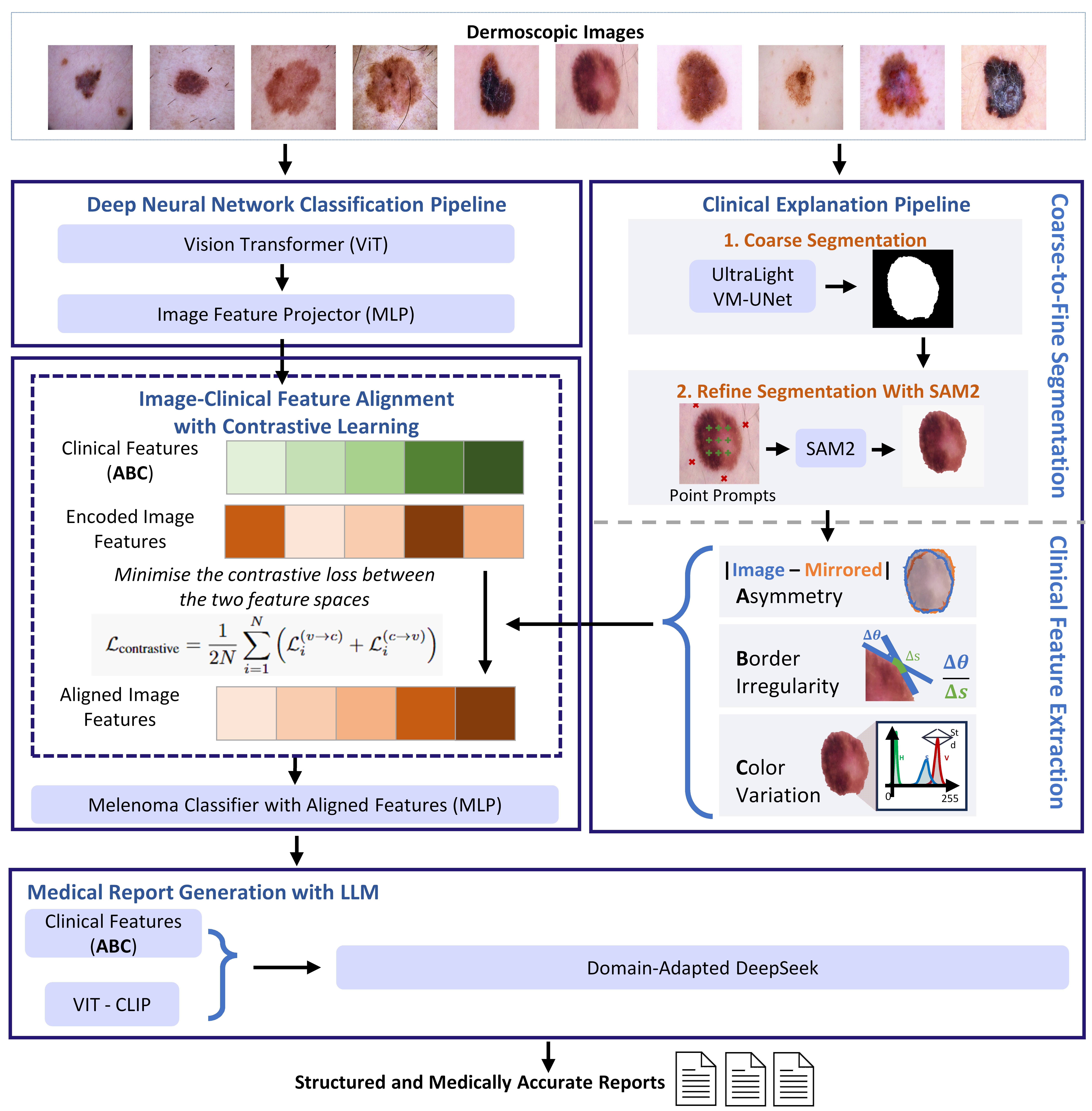}
    \caption{We propose a cross-modal melanoma diagnosis framework integrating four interconnected pipelines: a ViT-based classification pipeline that extracts semantic image features, a clinical explanation pipeline that segments lesions and quantifies ABC criteria, a contrastive module that aligns visual and clinical representations for interpretability, and a report generation module that produces structured diagnostic reports using CLIP descriptors and a domain-adapted LLM (DeepSeek).}
    \label{fig:overview}
\end{figure*}

\section{Methodology}
Our proposed \textbf{Cross-modal Explainable Framework for Melanoma diagnosis (CEFM)} is designed to integrate accurate image-based classification with clinically meaningful explanations, enhancing the interpretability of image classification models while preserving high performance. As illustrated in Figure~\ref{fig:overview}, CEFM comprises four components: 
\begin{enumerate}
    \item a \textit{Deep Neural Network (DNN) Classification Pipeline}, 
    \item a \textit{Clinical Explanation Pipeline}, 
    \item a \textit{Contrastive Learning Module}, and
    \item a \textit{Medical Report Generation Module}. 
\end{enumerate}    
The DNN Classification Pipeline extracts semantic image embeddings from dermoscopic images using a VIT, followed by an image feature projector for dimensional alignment. In parallel, the Clinical Explanation Pipeline performs coarse-to-fine lesion segmentation using UltraLight VM-UNet and SAM2, from which three clinically relevant features—\textit{Asymmetry}, \textit{Border Irregularity}, and \textit{Color Variation}—are computed. These image and clinical features are then fused via a cross-modal contrastive learning mechanism that encourages alignment between visual embeddings and diagnostic priors, thereby enhancing interpretability. Finally, to facilitate clinician understanding and trust, CEFM includes a report generation module that leverages CLIP-based concept activation and a domain-adapted large language model to produce structured and medically grounded diagnostic reports. We introduce them as follows.

\begin{table*}[ht]
    \small
    \centering
    \begin{tabular}{|c|c|c|c|}
        \hline
        \textbf{Feature} & \textbf{Clinical Concept} & \textbf{Calculation Formula} & \textbf{Interpretation} \\
        \hline
        \hline
        Asymmetry (A) & \begin{tabular}{@{}c@{}}The two halves of the\\mole are asymmetrical\end{tabular} 
                      & $\displaystyle A = \frac{\sum |I(x,y) - I_{\text{mirror}}(x,y)|} {\sum M(x,y)}$ 
                      & \begin{tabular}{@{}c@{}}Differential pixels in lesion area / \\Total lesion area pixels\end{tabular} \\
        \hline
        Border (B)   & \begin{tabular}{@{}c@{}}The border of the mole\\is irregular, jagged, or blurred\end{tabular} 
                     & \begin{tabular}{@{}c@{}}$\displaystyle B_2 = \frac{1}{N} \sum_{i=1}^{N} \kappa_i$ where
                        $\displaystyle \kappa_i = \frac{\Delta \theta_i}{\Delta s_i}$\end{tabular}
                     & \begin{tabular}{@{}c@{}}Mean curvature of the lesion border\\ with $\kappa_i$ representing local curvature\\ at point $i$\end{tabular} \\
        \hline
        Color (C) & \begin{tabular}{@{}c@{}}The mole has uneven colour, \\which may include brown, black, and \\ sometimes pink, white, or blue spots\end{tabular} 
                  & 
        \begin{tabular}{@{}c@{}}$\sigma_H = \sqrt{\frac{1}{N}\sum_{i=1}^{N}(H_i - \mu_H)^2}$\\
        $\sigma_S = \sqrt{\frac{1}{N}\sum_{i=1}^{N}(S_i - \mu_S)^2}$\\
        $\sigma_V = \sqrt{\frac{1}{N}\sum_{i=1}^{N}(V_i - \mu_V)^2}$\end{tabular} & \begin{tabular}{@{}c@{}}Standard deviations of HSV\\ color channels\end{tabular} \\
        \hline
    \end{tabular}
\caption{Quantitative computation of ABC features from dermoscopic images. $I(x,y)$ and $I_{\text{mirror}}(x,y)$ are pixel values in the original and mirrored images, $M(x,y)$ is the lesion mask; $\kappa_i$ is the local curvature calculated from the angle change ($\Delta\theta_i$) and arc length ($\Delta s_i$), \textit{N} is the number of valid curvature sample points; $H_i$, $S_i$, and $V_i$ are the hue, saturation, and value; $\mu_H$, $\mu_S$, and $\mu_V$ are their respective mean values.}
\label{tab:abc_features}
\end{table*}

\subsection{Deep Neural Network Classification Pipeline}
The \textbf{DNN Classification Pipeline} in CEFM is responsible for extracting high-level semantic features from dermoscopic images to support melanoma classification and enable alignment with clinically derived features. We evaluated several backbone architectures for latent feature extraction, including ResNet50, DenseNet121, Inception-V3, EfficientNet-B2, and VGG16. Among these, the \textbf{ViT} consistently achieved the best classification performance and was therefore selected as the core feature extractor for our framework. 

Given an input dermoscopic image, the ViT encodes spatial and texture information into a fixed-dimensional latent representation. These representations are then passed through a \textbf{Multilayer Perceptron (MLP)} head, which acts as a feature projector to transform the ViT-encoded image embeddings into a common space suitable for contrastive alignment with clinical features. This MLP consists of two fully connected layers with non-linear activations, followed by $L_2$ normalization to ensure consistent feature scales.

The projected image embeddings serve dual purposes: they are used both for melanoma classification through a separate classification head and as input to the contrastive learning module, which aligns these features with clinically extracted ABC features. This modular design ensures that the learned representations are both discriminative and semantically aligned with dermatological reasoning.

\subsection{Clinical Explanation Pipeline}
The \textbf{Clinical Explanation Pipeline} is designed to extract human-interpretable, clinically relevant features from dermoscopic images, enabling transparent reasoning aligned with dermatological diagnostic practices. This pipeline complements the DNN-based classification by introducing domain knowledge through a two-stage process: \textit{lesion segmentation} and \textit{clinical feature extraction}. First, a coarse-to-fine segmentation module isolates the lesion region from surrounding skin using lightweight and accurate deep segmentation models. Then, clinically meaningful features—specifically \textbf{Asymmetry}, \textbf{Border Irregularity}, and \textbf{Color Variation}—are computed from the segmented lesion mask. These features correspond to the ``ABC'' components of the ABCD rule and serve as structured diagnostic priors. The extracted features are later aligned with image representations computed in the DNN Classification Pipeline to ensure the {\bf internal consistency} between the two pipelines.

\paragraph{Lesion Segmentation}
Lesion segmentation serves as the foundational step in the Clinical Explanation Pipeline, aiming to accurately delineate the melanoma region from the surrounding skin. This ensures that all downstream computations are based on the relevant lesion area. To achieve this, we employ a two-stage segmentation strategy for accurate lesion masks with minimal manual intervention. Coarse segmentation uses a pretrained UltraLight VM-UNet (trained on ISIC 2018), efficiently delineating lesion regions but yielding rough boundaries. These coarse masks serve as pseudo‑labels for subsequent refinement. In fine segmentation, we leverage SAM \cite{ravi2024sam}: foreground and background points are sampled automatically from the coarse masks to prompt SAM2, which generates multiple candidates per image; we then select the mask with highest IoU against the pseudo‑label. This yields high‑precision lesion contours for downstream ABC feature computation.

\paragraph{Clinical Feature Extraction}
From the segmented lesion images, we designed computational functions to automatically extract clinical features that are routinely used in dermatological practice to assess the malignancy risk of skin lesions.  Our analysis focuses on three criteria that can be reliably and robustly computed: \emph{Asymmetry}, \emph{Border}, and \emph{Colour}. The \emph{Differential structure} feature is excluded due to the lack of fine-grained expert annotations, as the identification and delineation of these structures typically require precise, region-specific labeling by dermatology specialists. For each of the retained criteria, we design quantitative feature descriptors derived from the refined segmentation masks (Table~\ref{tab:abc_features}). 

\begin{figure}[!ht]
    \centering
    \includegraphics[width=\linewidth]{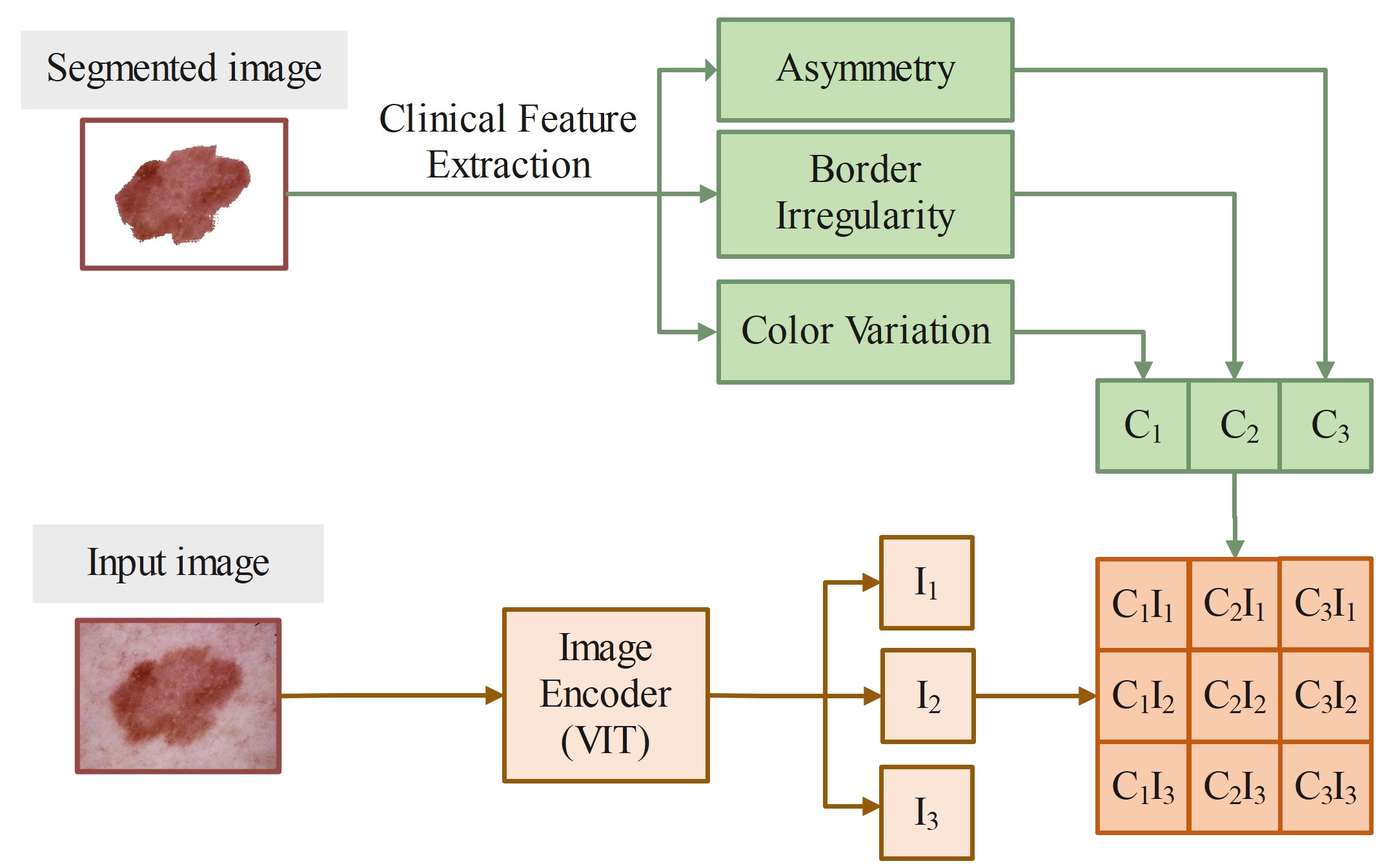}
    \caption{Overview of the cross-modal alignment process. Clinical features extracted from lesion segmentations are projected into a shared latent space, while image features from a ViT encoder are jointly aligned via element-wise interactions to supervise contrastive learning.}
    \label{fig:Contrastive Learning structure}
\end{figure}

\subsection{Feature Spaces Alignment with Contrastive Learning}
To bridge the semantic gap between deep image features and clinically meaningful descriptors, we introduce a \textbf{cross-modal contrastive learning} approach that aligns high-level visual representations, computed in the DNN Classification Pipeline, with structured clinical features computed in the Clinical Explanation Pipeline. This alignment grounds abstract image embeddings in interpretable dermatological concepts, enhancing both model transparency and robustness. An overview of the cross-modal interaction is illustrated in Figure~\ref{fig:Contrastive Learning structure}.

Inspired by SimCLR~\cite{chen2020simple} and BYOL~\cite{grill2020bootstrap}, we construct two modality-specific branches for contrastive learning—one for dermoscopic image features and another for clinical feature vectors. Each branch contains a projection head that maps modality-specific embeddings into a shared latent space where semantic alignment is enforced through contrastive objectives.
Formally, let $x$ denote an input image and $c$ the corresponding clinical feature vector. A pretrained image encoder $f_v(\cdot)$ extracts visual features $v = f_v(x) \in \mathbb{R}^{d_v}$, while a lightweight MLP $f_c(\cdot)$ encodes the clinical features as $u = f_c(c) \in \mathbb{R}^{d_c}$. These embeddings are then projected into a common embedding space using two multilayer perceptrons (MLPs), $h_v$ and $h_c$:
\begin{equation}
z_v = h_v(v), \quad z_c = h_c(u), \quad z_v, z_c \in \mathbb{R}^d.
\end{equation}
To facilitate contrastive learning based on cosine similarity, we apply $\ell_2$ normalization:
\begin{equation}
\tilde{z}_v = \frac{z_v}{\|z_v\|_2}, \quad \tilde{z}_c = \frac{z_c}{\|z_c\|_2}.
\end{equation}
We optimize the alignment using normalized temperature-scaled cross-entropy loss (NT-Xent). The loss for a positive image-clinical pair $(\tilde{z}_{v_i}, \tilde{z}_{c_i})$ is defined as:
\begin{equation}
\mathcal{L}_i = - \log \frac{\exp(\text{sim}(\tilde{z}_{v_i}, \tilde{z}_{c_i}) / \tau)}{\sum_{j=1}^{N} \exp(\text{sim}(\tilde{z}_{v_i}, \tilde{z}_{c_j}) / \tau)},
\end{equation}
where $\text{sim}(a, b) = a^\top b$ is the cosine similarity and $\tau$ is the temperature hyperparameter. We compute a bidirectional contrastive loss by averaging over the batch:
\begin{equation}
\mathcal{L}_{\text{contrastive}} = \frac{1}{2N} \sum_{i=1}^{N} \left( \mathcal{L}_i^{(v \rightarrow c)} + \mathcal{L}_i^{(c \rightarrow v)} \right).
\end{equation}

During training, we freeze the parameters of the pretrained image encoder and optimize only the projection heads, preserving the semantic structure of visual features while enforcing cross-modal alignment. 
After contrastive pretraining, the projection head is frozen, and a lightweight classification head is trained on the projected image embeddings for downstream melanoma classification. This design enables clinically grounded visual representations that support both accurate prediction and meaningful explanation.


\begin{table*}[t]
    \centering
    \begin{tabular}{|c|p{14.75cm}|}
        \hline
        Lesion Type & CLIP-Matched Clinical Feature \\
        \hline
        Benign Lesion & 
        Regular symmetric shape, Smooth borders, Uniform color, Light brown color, Small size, Well-defined edges, No ulceration, No bleeding, No satellite lesions, Dome-shaped structure, Single color tone \\
        \hline
        Melanoma & 
        Asymmetric shape, Irregular borders, Uneven color, Dark brown areas, Black areas, Blue-gray areas, Red areas, White areas, Multiple mixed colors, Fuzzy edges, Ulceration, Bleeding, Crusting, Scaling, Satellite lesions \\
        \hline
    \end{tabular}
    \caption{CLIP-matched dermatological attributes used for interpretable lesion classification. These attributes are curated to align with clinical features of benign nevi and malignant melanoma for visual-language modeling.}
    \label{tab:clip_attributes}
\end{table*}

\begin{table*}[t]
\centering
\begin{tabular}{|>{\centering\arraybackslash}p{2cm}|p{14.75cm}|}
\hline
{\bf AI Diagnosis} & \multicolumn{1}{|c|}{\textbf{Suspicious for Melanoma}} \\
\hline
\hline
Quantitative  & \textbf{Asymmetry (A)}: 11.69 – severe asymmetry, strong malignancy indicator\\
Features      & \textbf{Border Irregularity (B2)}:2.71 – highly irregular and serrated borders\\
              & Color Variation (H): 1.02 – minimal hue variation; Saturation Variation (S): 2.17 – mild; Brightness Variation (V): 2.53 – mild\\
\hline
\multirow{2}{2cm}{\centering Visual Features} & Satellite lesions – indicative of possible metastasis\\
                & \textbf{Ulceration} – high-risk prognostic feature\\
                & Dark brown pigmentation – associated with melanocytic proliferation\\
\hline
\multirow{2}{2cm}{\centering Risk Assessment} & High suspicion due to: \textbf{severe asymmetry, border irregularity, satellite lesions, ulceration.} \\
                & Low hue variation does not exclude malignancy. \\
\hline
\end{tabular}
\caption{Excerpt from AI-generated diagnostic report. Bolded features indicate high-risk indicators.}
\label{tab:ai_report_sample}
\end{table*}

\subsection{Report Generation with Large Language Model}
To improve interpretability and clinical usability, we introduce a medical report generation module that translates model outputs into structured, natural language summaries. We first discretize the computed ABC features into five severity levels. In parallel, we use CLIP-ViT-B/16 to retrieve top-ranked melanoma-related descriptors (e.g., ``asymmetric shape,'' ``blue-gray areas'') based on image features. CLIP is a vision-language model trained via contrastive learning to align image and text embeddings within a shared latent space, enabling the retrieval of semantically meaningful textual descriptions by computing similarity scores in this multimodal space. These quantitative and semantic cues are then combined into prompts and fed into DeepSeek, a domain-adapted large language model, to generate concise, medically grounded diagnostic reports. Table~\ref{tab:clip_attributes} summarizes the CLIP descriptors, and Table~\ref{tab:ai_report_sample} shows a representative report.

\section{Model Performance Quantitative Evaluation}
\label{sec:results}

\paragraph{Datasets}
To develop and evaluate our framework, we curated two benchmark dermoscopic datasets from the International Skin Imaging Collaboration (ISIC) Archive—the largest publicly available repository of quality-controlled skin lesion images. We used the ISIC 2018 dataset to train and evaluate the coarse segmentation model, as it provides expert-annotated lesion masks. For the classification and fine segmentation tasks, we selected 3,130 high‑quality melanoma and nevus dermoscopic images from ISIC 2020 (excluding cases with ambiguous boundaries or severe artifacts) and supplemented them with 448 unique melanoma images from ISIC 2018 to mitigate class imbalance—resulting in 3,578 images (646 melanoma, 2,932 nevi) for training—and evaluated performance on the official ISIC 2020 validation (n = 573) and test (n = 714) splits.


\paragraph{Segmentation Performance}
As the first stage in our segmentation pipeline, we employ the UltraLight VM-UNet—an existing lightweight architecture—as the backbone for coarse lesion mask generation. To validate its suitability for generating reliable pseudo-labels, we benchmarked its segmentation performance on the ISIC2018 dataset under consistent evaluation settings with several representative classical and lightweight models. 
UltraLight VM-UNet achieves a Dice Similarity Coefficient (DSC) of 0.8909, accuracy of 95.56\%, and specificity of 0.9746, outperforming most baselines in overall performance.

\paragraph{Comparison of Vision Models on ISIC2020}
ViT and EfficientNet-B2 achieve the highest accuracy (94.26\%), with ViT further exhibiting the best precision (88.19\%), AUC (0.972), and specificity (97.15\%). Given its strong overall performance and suitability for contrastive learning frameworks like CLIP, ViT is adopted as the baseline for subsequent experiments.

\paragraph{Classification Performance after Feature Alignment}
To evaluate the effectiveness of our cross-modal contrastive learning framework, we froze the image encoder and projection head, then fine-tuned lightweight classification heads on their embeddings. 
All experiments were repeated 10 times, with performance reported as mean\,$\pm$\,standard deviation. The ViT-based model achieved the highest overall performance, with an accuracy of $92.79\%\pm0.57\%$ and an AUC of $0.961\pm0.004$.
Given its high performance and architectural compatibility with the CLIP-based explainability module, ViT is selected as the backbone of our diagnostic pipeline.


\begin{figure}[!ht]
    \centering
    \includegraphics[width=\linewidth]{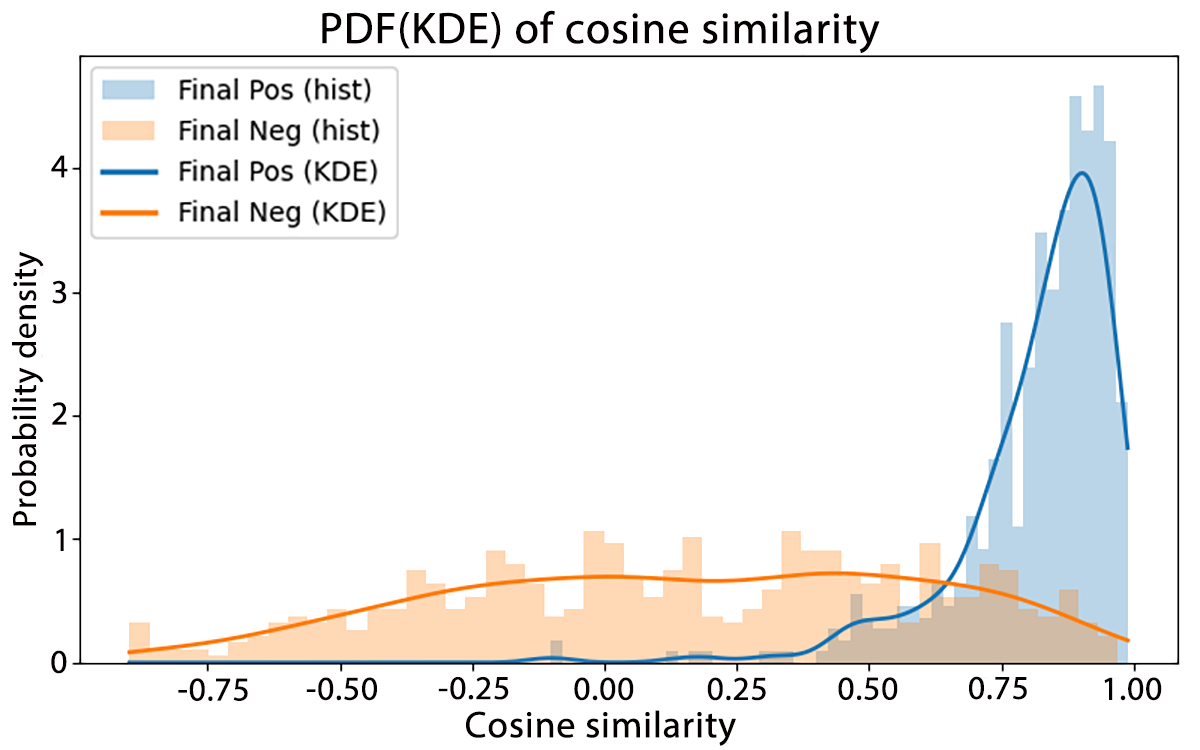}
    \caption{The figure shows cosine similarity distributions of positive and negative pairs after contrastive learning. Positive pairs cluster in the high-similarity region ($>0.75$), whereas negative pairs occupy lower values, indicating effective cross-modal alignment.}
    \label{fig:cosine}
\end{figure}

\paragraph{Feature Alignment via Contrastive Learning}
\label{sec:tsne}
We evaluate feature alignment by analyzing the cosine similarity distributions between paired images and clinical embeddings after contrastive learning. As shown in Fig.~\ref{fig:cosine}, positive pairs exhibit significantly higher similarity values, concentrated near 0.8, while negative pairs are distributed near zero. This clear separation demonstrates that contrastive learning effectively aligns cross-modal features and improves discriminative capability in the shared embedding space.

\begin{figure*}[ht]
    \centering
    \includegraphics[trim={0 8.4074cm 0 0.3556cm},clip, width=\linewidth]{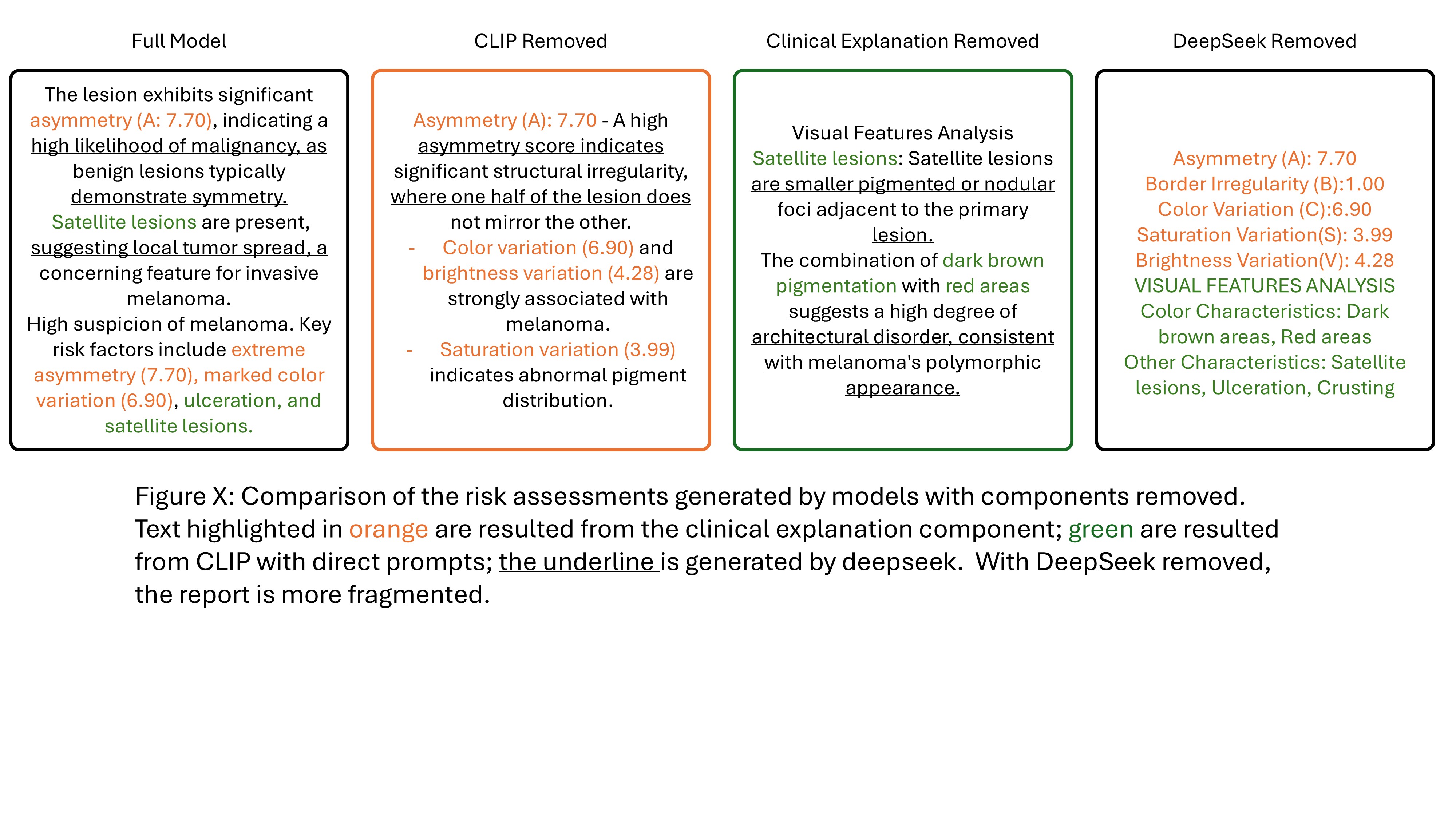}
    \caption{Comparison of the risk assessments generated by models with components removed. Text highlighted in orange are resulted from the clinical explanation component; green are resulted from CLIP with direct prompts; the underline is generated by deepseek.  With DeepSeek removed, the report is more fragmented.}
    \label{fig:Ablation experiment}
\end{figure*}

\begin{table*}[ht]
\centering
\begin{tabular}{|p{6.5cm}|c|c|c|c|}
\hline
\textbf{Evaluation Dimension} & \textbf{Expert 1} & \textbf{Expert 2} & \textbf{Expert 3}  & \textbf{Average Score} \\
\hline
1. Consistency with clinical judgment           & 4.55 & 4.18 & 3.81 & 4.18\\
2. Usefulness of ABC feature analysis           & 5.00 & 4.20 & 4.00 & 4.40 \\
3. Report clarity and readability               & 4.50  & 4.50 & 4.50 & 4.50\\
4. Interpretability of AI decision-making       & 5.00 & 4.40 & 4.40 & 4.60\\
5. Clinical applicability and decision support  & 4.50 & 3.50 & 4.00 & 4.00 \\
\hline
\end{tabular}
\caption{Quantitative Expert Evaluation (Likert scale: 1 = strongly disagree; 5 = strongly agree)}
\label{tab:expert-score}
\end{table*}

\paragraph{LLM Report Comparison}
We constructed a single prompt to generate dermatology reports---structured into quantitative feature analysis, visual feature description, consistency and complementarity evaluation, and risk assessment with treatment recommendations---and applied it to \textbf{Anthropic Claude}, \textbf{Google Gemini Base}, and \textbf{DeepSeek}. We then used BERTScore (RoBERTa-Large, layer~9) to compare DeepSeek’s outputs against those of Claude and Gemini, reporting Precision, Recall, and F1 to quantify semantic similarity.  Given the comparable semantic performance across models and DeepSeek’s superior cost-effectiveness, we selected DeepSeek for subsequent analyses.

\paragraph{Ablation Study}
To assess the contribution of each component in our explanation pipeline, we conducted an ablation study comparing outputs with individual modules removed (Figure~\ref{fig:Ablation experiment}). The \textbf{CLIP module} supplements the report with additional visual cues that complement the quantified ABC features. When CLIP is removed, the report centers on ABC features and their numerical values, lacking additional contextual factors that may affect the diagnosis. Removing the \textbf{clinical explanation module} eliminates critical quantification of ABC features, resulting in text that relies solely on visual descriptions without numerical interpretation. The \textbf{DeepSeek module} plays a central role in synthesizing these elements into coherent and well-structured narrative reports, ensuring both fluency and contextual integration. Without DeepSeek, the report becomes fragmented, as it lacks the structured presentation that supports narrative coherence. These results underscore the complementary roles of each module in generating comprehensive, interpretable, and clinically meaningful melanoma risk assessments.

\section{User Study with Expert Dermatologists}
To comprehensively evaluate the interpretability, usability, and clinical applicability of our proposed multimodal explainable melanoma diagnostic framework, we conducted a structured user study involving three board-certified dermatologists with extensive clinical experience.
Each expert independently reviewed one system-level report and five lesion-level reports. For the system-level report, a broad questionnaire evaluated presentation clarity, AI model trustworthiness, clinical impact, and application. The questionnaire consists of Likert-scale questions assessing each dimension, followed by open-ended and multiple-choice questions to explore rationale, limitations, and applications.  
For the lesion-level reports, a focused questionnaire with Likert-scale and open questions assessed model accuracy and interpretability. 
Quantitative responses will be summarized using descriptive statistics (means) across the six evaluated reports. For qualitative responses, thematic analysis will be conducted using an inductive coding approach.
Table~\ref{tab:expert-score} summarizes expert scores.
Experts confirmed that structured ABC quantification offers clinically relevant and interpretable outputs consistent with physician intuition, establishing ABC reference ranges and diagnostic formats to complete the ABCD framework. They noted that its diagnostic consistency, transparency, and utility can enhance clinician trust and support decision-making.


\section{Conclusion}
We introduce \textbf{CEFM}, a cross‑modal framework for explainable melanoma diagnosis that integrates ViT‑based classification with clinically grounded ABC feature extraction and contrastive alignment. A CLIP‑driven LLM report module produces structured, transparent diagnostic summaries. Expert evaluation confirms the system’s interpretability and clinical utility, yielding a mean rating of 4.6/5. Additionally, Our experiments on the ISIC dataset demonstrate strong performance, with 92.79 \% accuracy and an AUC of 0.961. 

Despite its promising results, CEFM has two limitations. First, the exclusion of the \textbf{differential structure} criteria from the ABCD rule—due to the lack of expert-level annotated data—limits the completeness of assessment. Second, the current system assumes high-quality dermoscopic inputs and may underperform in cases involving poor imaging conditions or rare lesion subtypes not represented in the training data. Future work will integrate differential structure analysis, support multi‑temporal lesion tracking, and validate the framework in prospective clinical studies to enhance robustness across diverse lesion types and imaging conditions.

\section{Acknowledgements}
This research was supported by the Ministry of Education, Singapore (Grant ID: RS15/23).

\bibliography{aaai2026}

\end{document}